# Transformer-Based Multi-Modal Temporal Embeddings for Explainable Metabolic Phenotyping in Type 1 Diabetes

Pir Bakhsh Khokhar
*Department of Computer Science*
*University of Salerno*
Salerno, Italy
pkhokhar@unisa.it

Carmine Gravino
*Department of Computer Science*
*University of Salerno*
Salerno, Italy
gravino@unisa.it

Fabio Palomba
*Department of Computer Science*
*University of Salerno*
Salerno, Italy
fpalomba@unisa.it

Sule Yildirim Yayilgan
*Department of Info. Security and Communication Tech.*
*Norwegian University of Science and Technology*
Gjøvik, Norway
sule.yildirim@ntnu.no

Sarang Shaikh
*Department of Info. Security and Communication Tech.*
*Norwegian University of Science and Technology*
Gjøvik, Norway
sarang.shaikh@ntnu.no

*Abstract*—Type 1 diabetes (T1D) is a highly metabolically heterogeneous disease that cannot be adequately characterized by conventional biomarkers such as glycated hemoglobin (HbA1c). This study proposes an explainable deep learning framework that integrates continuous glucose monitoring (CGM) data with laboratory profiles to learn multimodal temporal embeddings of individual metabolic status. Temporal dependencies across modalities are modeled using a transformer encoder, while latent metabolic phenotypes are identified via Gaussian mixture modeling. Model interpretability is achieved through transformer attention visualization and SHAP-based feature attribution. Five latent metabolic phenotypes, ranging from metabolic stability to elevated cardiometabolic risk, were identified among 577 individuals with T1D. These phenotypes exhibit distinct biochemical profiles, including differences in glycemic control, lipid metabolism, renal markers, and thyrotropin (TSH) levels. Attention analysis highlights glucose variability as a dominant temporal factor, while SHAP analysis identifies HbA1c, triglycerides, cholesterol, creatinine, and TSH as key contributors to phenotype differentiation. Phenotype membership shows statistically significant, albeit modest, associations with hypertension, myocardial infarction, and heart failure. Overall, this explainable multimodal temporal embedding framework reveals physiologically coherent metabolic subgroups in T1D and supports risk stratification beyond single biomarkers.

## I. INTRODUCTION

Type 1 diabetes (T1D) is an autoimmune disease characterized by the progressive loss of pancreatic $\beta$-cells, resulting in lifelong insulin dependence and pronounced metabolic variability. Despite advances in insulin therapy and continuous glucose monitoring (CGM), substantial inter-individual differences in glycemic patterns and metabolic responses persist. Episodic hyperglycemia and increased glycemic variability contribute to microvascular and macrovascular complications, including nephropathy, retinopathy, and cardiovascular disease [1], [2]. The conventional biomarker glycated hemoglobin (HbA1c) reflects average glucose exposure but fails to capture rapid glucose fluctuations and short-term metabolic dynamics [3]. As a result, patients with similar HbA1c levels may exhibit markedly different risk profiles and comorbidity burdens, underscoring the need for a more granular characterization of metabolic heterogeneity in T1D [4], [5].

Machine learning (ML) methods have enabled data-driven disease subtyping using static clinical and laboratory variables [4], [6]. However, most approaches do not account for the dynamic nature of glucose metabolism. CGM signals are nonlinear and irregular, limiting the applicability of traditional clustering and regression methods to high-resolution glucose time series [7]. Transformer models, originally developed for natural language processing, can model long-range dependencies and cross-modal relationships, making them suitable for longitudinal metabolic data. Combined with biochemical measurements, transformers can learn multimodal temporal embeddings that provide compact and information-rich representations of patient-specific metabolic behavior.

A key limitation of deep temporal models is interpretability and clinical plausibility [8], [9]. In clinical settings, black-box predictions hinder adoption because trust and transparency are required. Explainability methods (XAI), including SHapley additive exPlanations (SHAP) and attention visualization, have shown value in uncovering model rationale and identifying physiologically relevant features [10], [11].

However, few studies have applied explainability to unsupervised temporal embeddings, particularly in metabolic diseases such as T1D, where time-dependent and biochemical variables interact [12], [13]. Moreover, links between latent representations and real-world outcomes, including cardiovascular and endocrine comorbidities, remain insufficiently explored.

To address these gaps, we propose an explainable multimodal temporal embedding model to identify latent metabolic phenotypes in T1D. The model integrates CGM trajectories with biochemical indicators, including HbA1c, triglycerides, total and high-density lipoprotein (HDL) cholesterol, creatinine, and thyrotropin (TSH), to generate unified patient-level embeddings. A transformer encoder captures short- and long-term cross-modal temporal dependencies, while Gaussian Mixture Modeling (GMM) identifies latent phenotypes in the learned representation space. Interpretation is achieved through attention visualization, highlighting influential glucose segments, and SHAP-based feature attribution, quantifying the contribution of each biomarker to phenotype separation.

Beyond unsupervised discovery, phenotypes were associated with diagnostic outcomes coded by the International Classification of Diseases (ICD-9/10), including hypertension, myocardial infarction, and heart failure. In a cohort of 577 patients with T1D, five physiologically interpretable metabolic phenotypes were identified, ranging from metabolic stability to elevated cardiometabolic risk. Lower-risk phenotypes exhibited more stable glucose profiles and higher TSH levels, whereas higher-risk phenotypes showed elevated HbA1c, triglycerides, and cholesterol.

SHAP analysis identified HbA1c, triglycerides, cholesterol, creatinine, and TSH as the most discriminative variables, and outcome analysis revealed significant associations between phenotypes and disease risk.

Overall, this study makes three contributions:
- A transformer-based framework for joint temporal embedding of CGM and biochemical data, capturing short- and long-term metabolic dynamics.
- An explainable pipeline combining attention visualization and SHAP-based attribution for transparent phenotype interpretation.
- A clinically aligned phenotyping approach linking data-driven phenotypes to cardiovascular and endocrine outcomes.

These contributions establish a foundation for explainable precision diabetology and the development of transparent digital biomarkers for individualized diabetes management.

## II. RELATED WORK

The processing of medical time-series data has been advanced by transformer-based architectures that can capture long-range dependencies and context-specific relationships across heterogeneous modalities. In the context of T1D, recent research has emphasized the need for models capable of explaining latent metabolic phenotypes, which static biomarkers cannot describe. The following briefly presents recent developments in transformer models, multimodal temporal embedding, and explainable deep phenotyping.

Early studies established transformers as strong alternatives to recurrent networks for learning clinical time-series representations [14]–[16]. The self-attention mechanism efficiently models complex temporal interactions and irregular sampling patterns common in biomedical data. Transformer-based models have outperformed other approaches in patient trajectory modeling, disease progression prediction, and individualized risk assessment when applied to electronic health records (EHRs) and physiological signals [17], [18]. Importantly, they scale effectively to integrate multimodal data, such as laboratory results, medication history, and clinical notes, producing unified patient-level embeddings [13], [19], [20].

The conceptual foundation of these embedding schemes follows the latent-space hypothesis, which proposes that patient trajectories exist in a continuous latent space in which disease states or phenotypes emerge as discrete clusters. Mapping metabolic trajectories onto this manifold facilitates the discovery of hidden phenotypes and disease progression pathways. This concept has been validated in studies applying similar frameworks to multi-omics and longitudinal clinical data in type 2 diabetes [4], [6] and metabolic syndrome [7], [21]. However, its use in T1D, particularly in combination with high-frequency CGM data, remains limited.

Recent progress in multimodal transformers has enabled the combination of temporal and biochemical data. For example, the Cross-Modal Temporal Pattern Discovery (CTPD) model enables dynamic phenotype discovery from numeric time-series and text data [16]. Other architectures include BEHRT (BERT for EHRs) and MedBERT, which pre-train transformers on large-scale EHRs to produce transferable patient embeddings [19], [22]. More recent models, such as TimelyGPT and RETAIN-3 [23], [24], use hybrid attention mechanisms to model temporal dependencies in irregularly sampled biomedical sequences. These developments suggest that interpretable embeddings capable of representing metabolic variability and disease progression are feasible.

Explainability is a key requirement in deep clinical modeling. Early methods such as SHapley Additive exPlanations (SHAP) and Local Interpretable Model-agnostic Explanations (LIME) provided local feature attribution [8], [10]. However, attention weights alone are not sufficiently interpretable for medical use and require association with clinically meaningful temporal events [9], [11]. Recent reviews suggest that embedding explainability directly into model architectures can improve trust and clinical adoption [16], [25], [26]. Hybrid approaches combining transformers with causal or probabilistic reasoning have also been proposed to bridge representation learning and mechanistic interpretability [27], [28].

Despite these advances, several challenges remain. Many transformer-based biomedical models rely on static or aggregated data, neglecting the fine-grained temporal variability characteristic of T1D. Multimodal integration has largely focused on EHR data, with limited use of high-resolution CGM signals and biochemical measurements. Explainability in unsupervised temporal embeddings is also understudied, and few works link latent phenotypes to real-world outcomes such as cardiovascular or endocrine comorbidities. Overall, although transformer models and explainable embeddings have progressed in biomedical time-series analysis, their application to T1D phenotyping remains limited. Most prior studies focus on glucose prediction or trend analysis rather than interpretable latent phenotype discovery. The present study addresses these gaps by jointly modeling continuous glucose dynamics and laboratory biomarkers using explainable unsupervised transformer embeddings.

## III. MATERIALS AND METHODS

To design the proposed framework and its experimental validation, this study addressed the following research questions (RQs):

**RQ1:** Can temporal embeddings characterize metabolic variability in T1D using CGM and laboratory biomarkers?

**RQ2:** Can these embeddings reveal distinct and clinically meaningful metabolic phenotypes beyond conventional biomarkers such as HbA1c?

**RQ3:** How can temporal and biochemical determinants of phenotype differentiation be identified using explainability methods?

To answer these questions, we developed a transformer-based multimodal temporal embedding framework that integrates heterogeneous temporal and biochemical data to identify explainable metabolic phenotypes in T1D.

### A. Data Sources

We used the open-access longitudinal *T1DiabetesGranada* repository from the Clinical Unit of Endocrinology and Nutrition, San Cecilio University Hospital, Granada, Spain [29]. The dataset includes CGM, laboratory, and diagnostic data collected with ethical approval (protocol K134665CRL). The cohort comprised 736 individuals with T1D monitored between January 2018 and March 2022, totaling more than 257,000 patient-days and 22.6 million glucose measurements. Glucose was recorded every 15 minutes using FreeStyle Libre and Libre2 devices (Abbott Diabetes Care, USA).

Each patient record included:
- Interstitial glucose concentrations (40–500 mg/dL) with timestamps.
- Laboratory measurements: glycated hemoglobin (HbA1c), triglycerides, total and high-density lipoprotein (HDL) cholesterol, creatinine, thyrotropin (TSH), insulin, and uric acid.
- ICD-9/ICD-10-coded comorbidities: hypertension, hyperlipidemia, myocardial infarction, and heart failure.
- Demographic variables: age, sex, and study duration.

A subsample of 577 participants with complete CGM and biochemical data was retained to ensure alignment between multimodal time series and laboratory measurements.

### B. Data Preprocessing

CGM data were resampled at 5-minute intervals and segmented into 24-hour windows to ensure temporal comparability across participants [30], [31]. Gaps shorter than 30 minutes were linearly interpolated, while longer gaps were excluded to preserve data integrity [32]. Laboratory measurements were aligned by matching each participant's median CGM date with the corresponding laboratory test date [33]. Categorical variables were one-hot encoded, and continuous variables were z-score normalized [34]. Extreme glucose values outside physiologically plausible ranges were excluded following international CGM interpretation guidelines [35]. The final dataset comprised approximately 240,000 temporally aligned CGM–laboratory pairs.

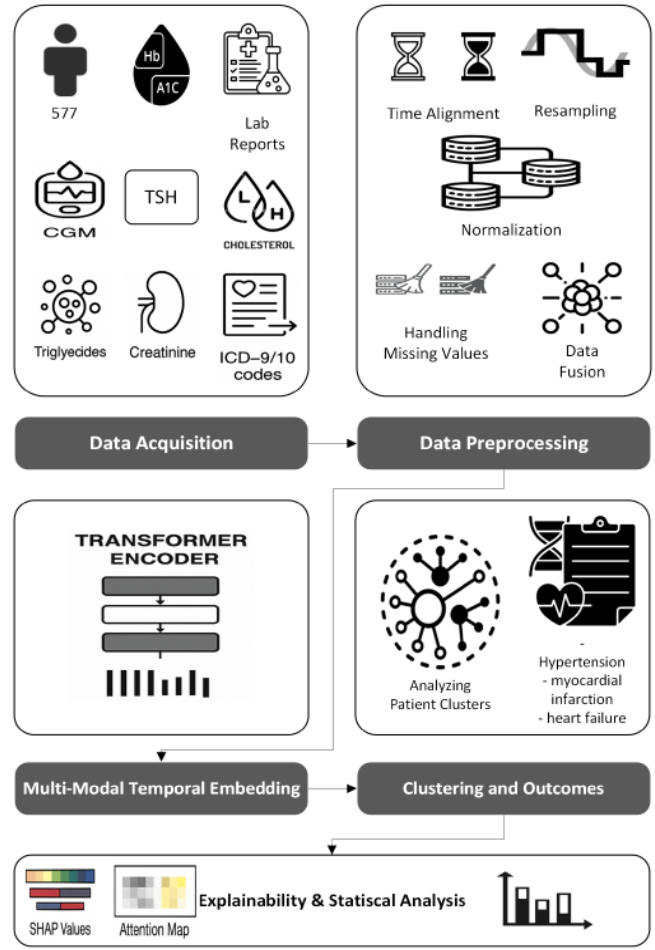

Fig. 1. Overview of the transformer-based multimodal temporal embedding framework for explainable metabolic phenotyping in type 1 diabetes.

### C. Transformer-Based Temporal Embeddings

A transformer encoder was used to extract patient-level embeddings by jointly modeling normalized CGM sequences and laboratory biomarkers as contextual tokens. The encoder employed four self-attention heads, 128-dimensional embeddings, and positional encodings to preserve temporal order. Training was unsupervised, using a masked reconstruction loss that minimized the mean squared error (MSE) between input and reconstructed glucose sequences. The resulting embeddings represent individualized temporal–biochemical trajectories. Transformers were selected for their ability to capture long-range temporal dependencies and integrate heterogeneous modalities [36]–[38]. Unlike recurrent architectures, which scale poorly for multimodal integration, this approach captures nonlinear cross-modal relationships [39], [40].

### D. Latent Phenotype Discovery

Latent metabolic phenotypes were identified using Gaussian Mixture Modeling (GMM) applied to the transformer embeddings (Fig. 1). GMM was chosen for its probabilistic assignment and support for soft cluster membership, which are appropriate for continuous and overlapping metabolic phenotypes. In contrast to hard clustering methods such as k-means, GMM captures gradual transitions between metabolic states. The optimal number of clusters was selected using the Bayesian Information Criterion (BIC) and silhouette analysis [41]. Reproducibility was assessed across random initializations, and temporal stability was evaluated using the Adjusted Rand Index (ARI) [42].

### E. Explainability Framework

Two complementary explainability strategies were employed. Attention-weight visualization identified CGM time segments that most influenced temporal encoding, highlighting periods of elevated glucose variability. SHapley Additive exPlanations (SHAP) quantified the contribution of each biochemical variable to phenotype differentiation. Cluster-level mean absolute SHAP values identified key biomarkers, including HbA1c, triglycerides, cholesterol, creatinine, and TSH.

### F. Statistical and Outcome Analysis

Inter-cluster differences in laboratory and CGM-derived variables were assessed using the Kruskal–Wallis test with Benjamini–Hochberg correction for multiple comparisons. Effect sizes were computed as Cohen's $d$ for continuous variables and Cramér's $V$ for categorical outcomes, with significance set at $p < 0.05$. Cramér's $V$ quantified the strength of association between latent phenotypes and clinical outcomes independently of sample size.

Chi-squared tests evaluated associations between clusters and clinical outcomes, and Uniform Manifold Approximation and Projection (UMAP) was used to visualize phenotype separability (Fig. 1).[1]

## IV. RESULTS

### A. Latent Embedding Structure and Disease Organization

The transformer encoder showed stable convergence, reaching a reconstruction error of approximately $1.2 \times 10^3$ by the 50th epoch. This value corresponds to the mean squared error computed on normalized glucose values and reflects convergence and embedding stability rather than clinical performance. The resulting 128-dimensional embeddings formed partially overlapping clusters in the latent space (Fig. 2), consistent with continuous metabolic heterogeneity rather than sharply separable disease states. This structure suggests that the learned representations capture meaningful temporal and biochemical variation, which is further examined through phenotypic characterization and outcome associations.

[1] All resources are available on GitHub.

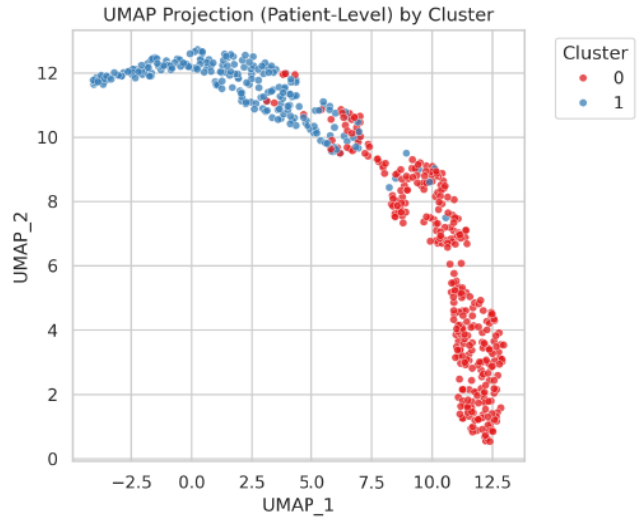

Fig. 2. UMAP projection of learned embeddings showing distinct patient clusters identified via Gaussian Mixture Modeling.

When embeddings were colored by major cardiometabolic outcomes, including diabetes mellitus, hypertension, hyperlipidemia, heart failure, and myocardial infarction, distinct regional enrichment was observed (Fig. 3). This indicates that the latent space preserves disease-relevant structure aligned with temporal–biochemical patterns.

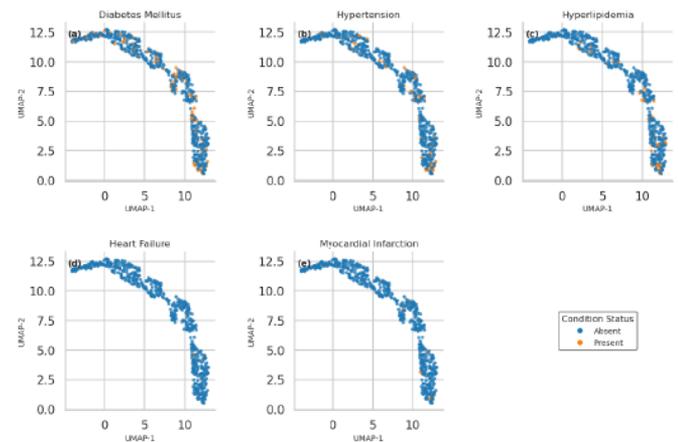

Fig. 3. Composite UMAP projections illustrating disease-specific organization of patient embeddings. Orange points denote patients diagnosed with each condition.

### B. Discovery of Latent Metabolic Phenotypes

Gaussian Mixture Modeling of the embeddings revealed five latent metabolic phenotypes. A silhouette score of 0.49 (Fig.

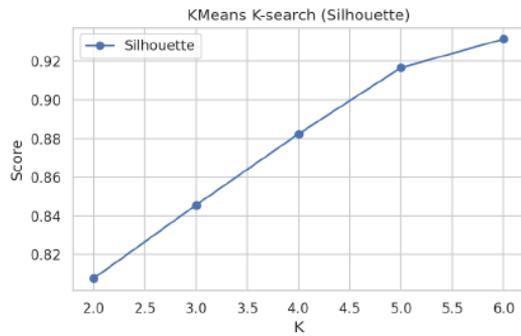

Fig. 4. Silhouette analysis confirming the optimal cluster number ($k = 5$) for GMM-based phenotype discovery.

4) supported five clusters as an optimal solution, reflecting well-defined yet continuous structure in the latent space.

Distinct biochemical profiles were observed across phenotypes (Fig. 5). Clusters 1 and 2 corresponded to metabolically stable or mildly dyslipidemic profiles; Cluster 3 reflected insulin resistance patterns; Cluster 4 exhibited multimorbidity; and Cluster 5 represented a high-risk cardiometabolic phenotype characterized by elevated HbA1c and triglycerides. These findings indicate that the model disentangles complex biochemical variability into physiologically meaningful subgroups.

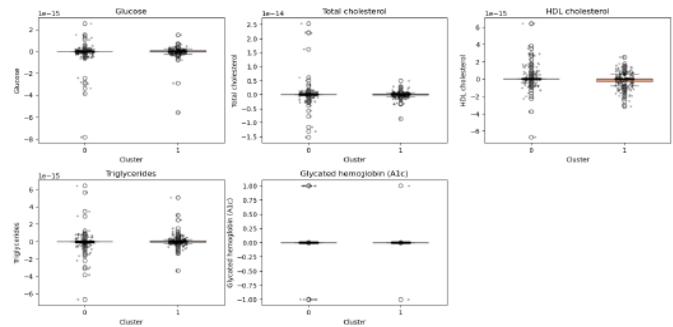

Fig. 5. Distribution of key biochemical variables (glucose, lipids, HbA1c) across latent phenotypes, highlighting metabolic differentiation.

### C. Explainable Feature and Temporal Attribution

Global SHapley Additive exPlanations (SHAP) identified lipid- and glycemic-related variables as the primary drivers of phenotype separation (Fig. 6). The strongest contributors were HDL cholesterol, total cholesterol, HbA1c, and creatinine, reflecting cardiometabolic and renal processes. Moderate contributions from thyrotropin (TSH) suggested additional endocrine involvement.

Attention heatmaps (Fig. 7) showed that the model focused on physiologically relevant glucose excursions, particularly postprandial spikes and rapid fluctuations. High-risk clusters exhibited broader and less localized attention patterns, consistent with unstable glucose regulation compared to metabolically stable phenotypes.

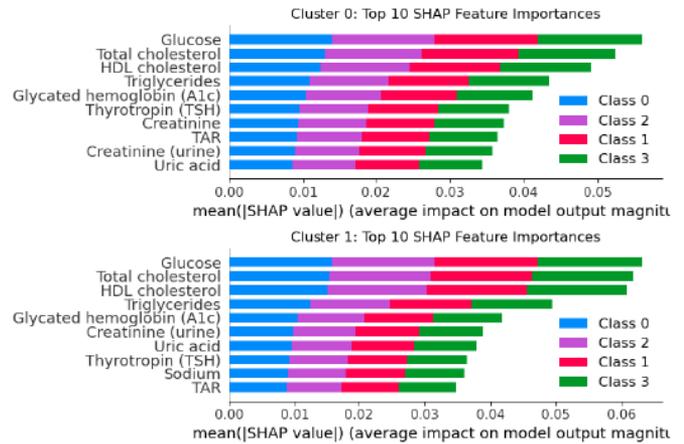

Fig. 6. Global SHAP feature importance showing dominant influence of lipid, glycemic, and renal markers in phenotype separation.

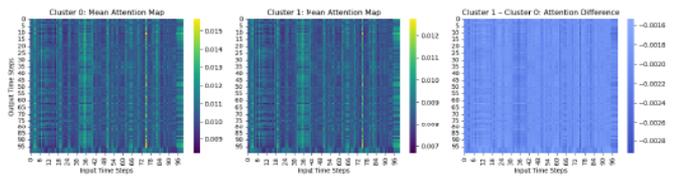

Fig. 7. Transformer attention maps highlighting temporal focus for Cluster 0 (stable), Cluster 1 (high-risk), and their difference. Regions of high attention correspond to periods of sharp glucose variation.

### D. Clinical Validation of Phenotypes

Associations between phenotype membership and clinical outcomes were quantified using Cramér's $V$ (Table I). The strongest associations were observed for hypertension, myocardial infarction, and heart failure, indicating distinct disease risk profiles across phenotypes.

TABLE I
CRAMÉR'S $V$ EFFECT SIZES BETWEEN LATENT PHENOTYPES AND MAJOR OUTCOMES.

| Outcome | Cramér's $V$ |
|---|---|
| Hypertension | 0.077 |
| Myocardial Infarction | 0.075 |
| Heart Failure | 0.072 |
| Sleep Disturbances | 0.063 |
| Carpal Tunnel Syndrome | 0.048 |

Outcome prevalence analysis (Table II) revealed a clear risk gradient, with Cluster 5 showing the highest prevalence of heart failure and myocardial infarction, and Cluster 1 exhibiting the lowest prevalence across outcomes. This trend aligns with the progression from metabolic stability to cardiometabolic deterioration observed in biochemical profiles.

Table II reports representative lower-risk and higher-risk phenotypes (Clusters 1 and 0, respectively), while the remaining clusters exhibit intermediate metabolic profiles.

TABLE II
PREVALENCE OF CARDIOMETABOLIC OUTCOMES (%) FOR
REPRESENTATIVE LOWER-RISK AND HIGHER-RISK PHENOTYPES.

| Cluster | DM | HF | HTN | HLD | MI |
|---|---|---|---|---|---|
| 0 | 16.8 | 0.8 | 11.0 | 8.1 | 1.7 |
| 1 | 14.4 | 0.0 | 6.6 | 6.5 | 0.3 |

Overall, the framework decomposed complex temporal–biochemical interactions into five clinically interpretable metabolic phenotypes. These phenotypes reflect progressive differences in glucose regulation, lipid metabolism, and renal function associated with increasing cardiometabolic risk. By integrating explainable clustering with transformer-based temporal embeddings, the model demonstrates both biological transparency and applicability to precision metabolic monitoring and patient stratification.

## V. DISCUSSION

*RQ1: Capturing Latent Metabolic Heterogeneity:*

Regarding **RQ1**, the transformer encoder learned latent representations that disentangled temporal changes in biochemical profiles. The model captured nonlinear interactions among glycemic, lipid, and renal biomarkers. Unlike static clustering methods, temporal embeddings represent patient trajectories, revealing heterogeneity consistent with real-world metabolic behavior rather than single-point measurements. Temporal–biochemical integration increases sensitivity to subtle metabolic changes that may precede obvious manifestation of the disease.

*RQ2: Explainability and Mechanistic Insight*

**RQ2** addressed whether the derived phenotypes were biologically interpretable, which identified lipid and glycemic markers, particularly HDL cholesterol, total cholesterol, HbA1c, and mean glucose, as the primary phenotype discriminators. Attention heatmaps showed that the model emphasized postprandial periods and high-variability glucose segments, corresponding to physiologically relevant episodes of metabolic stress. Together, these results demonstrate that the framework is both predictive and capable of clinically interpretable, mechanistically grounded attribution.

*RQ3: Clinical Relevance and Phenotypic Stratification*

**RQ3** The five identified clusters ranged from metabolically stable profiles to high-risk multimorbid groups characterized by insulin resistance, dyslipidemia, and elevated inflammatory burden, with strong associations to hypertension, myocardial infarction, and heart failure. These phenotypes are consistent with known patterns of cardiometabolic disease progression but refine overlapping pathophysiological axes beyond traditional diagnostic categories.

Several alternative approaches to metabolic phenotyping in diabetes have been proposed. Classical data-driven subtyping relies on static clinical and laboratory variables with conventional clustering, identifying coarse subgroups while neglecting dynamic glucose regulation [4], [6]. Other methods summarize CGM data using handcrafted metrics, such as time-in-range or variability indices, followed by linear dimensionality reduction or clustering, thereby discarding fine-grained temporal structure [3]. More recent approaches based on autoencoders or recurrent neural networks model temporal signals but have limited capacity to capture long-range dependencies and complex cross-modal interactions in high-resolution CGM data [13], [15]. In contrast, transformer-based temporal embeddings explicitly model long temporal contexts and multimodal relationships, making them well suited for unsupervised phenotyping of heterogeneous metabolic trajectories, as demonstrated in this study.

*Implications and Future Directions*

These findings highlight the value of explainable, time-resolved metabolic modeling at the patient level. By linking learned embeddings to interpretable biochemical drivers, this study bridges artificial intelligence and metabolic science. Future work should extend this framework to additional modalities, including genomics, lifestyle, and imaging data, and evaluate phenotype stability in larger prospective cohorts. Integrating causal inference or reinforcement learning could further enable dynamic risk prediction and adaptive therapeutic guidance in diabetes care.

*Limitations*

Laboratory measurements were unevenly distributed over time, and the cohort size and heterogeneity may limit generalizability. The identified phenotypes require validation in independent cohorts. Finally, although SHAP values and attention maps aid interpretation, they are correlational and should be complemented by causal or experimental analyses to confirm underlying mechanisms.

## VI. CONCLUSION

In this study, we proposed a model based on explainable transformers that learns temporal-biochemical embeddings to uncover hidden metabolic phenotypes for type 1 diabetes. The model recognized five clinically relevant profiles with a spectrum ranging from metabolic stability to high cardiometabolic risk. SHAP and attention analyses provided mechanistic transparency, with lipid and glycemic measures as key drivers, demonstrating a physiologically consistent temporal resolution. These phenotypes were highly correlated with important cardiometabolic outcomes and were therefore clinically significant. Overall, the framework promotes the use of interpretable deep learning for metabolic stratification and establishes a foundation for precision monitoring and risk assessment in cardiometabolic care.

## ACKNOWLEDGMENT

This work has been partially supported by the European Union through the Italian Ministry of University and Research, Project PNRR "D3-4Health: Digital Driven Diagnostics, prognostics and therapeutics for sustainable Health care". PNC 0000001. CUP B53C22006090001